# A standardised platform for translational advances in fluidic soft systems

Maks Gepner
Soft Systems Group, The School of Engineering
The University of Edinburgh
Edinburgh, Scotland
m.gepner@ed.ac.uk

Jonah Mack
Soft Systems Group, The School of Engineering
The University of Edinburgh
Edinburgh, Scotland
jmack@ed.ac.uk

Adam A. Stokes
Soft Systems Group, The School of Engineering
The University of Edinburgh
Edinburgh, Scotland
adam.stokes@ed.ac.uk

*Summary* — Soft machines are poised to deliver significant real-world impact, with soft robotics emerging as a key sub-discipline. This field integrates biological inspiration, materials science, and embodied intelligence to create bio-robotic hybrids, blurring the boundary between engineered systems and biology. Over the past 15 years, research in fluidically controlled soft robots has led to commercialised systems that leverage "softness" to improve human-machine interaction or to handle delicate objects. However, translating laboratory advancements into scalable applications remains challenging due to difficulties in prototyping and manufacturing ultra-flexible materials, as well as the absence of standardised design processes. Here we show that the Flex Printer, an open-source, low-cost FDM platform, enables reliable printing of ultra-flexible soft robots with embedded fluidic logic. By employing an innovative upside-down print orientation, the system significantly expands the range of printable geometries. We demonstrate how this approach allows robots to autonomously walk off the print bed immediately after fabrication—a milestone achievement in soft robotics. This work provides a foundation for standardisation and scalable manufacturing, critical for accelerating the field's impact. More broadly, by lowering barriers to entry, this platform has the potential to democratise soft robotics research and facilitate the development of new applications. We invite the community to contribute to the shared development of this technology to drive the next wave of breakthroughs in soft robotics.

*Index terms*—Soft robotics, 3D printing, Fluidics

## I. Introduction

*A. Soft machines are on the cusp of delivering outsized real-world impact.*

"Soft robotics" is a sub-discipline of a wider field of soft machines and soft systems [1]. Soft systems blend together advancements in three main areas: biological inspiration; materials science, and embodied/physical intelligence. Current cutting-edge research in soft systems is blurring the line between engineered systems and biology to create bio-robotic hybrids [2], [3].

Research activity in soft machines over the last 15 years, has shown huge progress in fluidically controlled soft robots and has led to commercialised systems (c.f. Soft Robotics Inc., Bioliberty Ltd., Imago Rehab. Ltd., Organic Robotics Company, Fluidic Logic Ltd., amongst many others) that use "softness" as a key engineering feature to solve problems in interfacing machines with humans or delicate objects.

The path to translating lab advancements to real-world applications has undoubtedly been slowed by a lack of standardised manufacturing and design processes, such as those that have been developed in for electronics and semiconductor manufacturing. In this paper, we introduce a fabrication platform and set of design rules that we believe will lay the groundwork for enabling facile translation of research developments between research labs, and into scaled-up manufacturing.

Research into fluidic soft machines has brought innovations that promise to greatly enhance human-robot collaboration [4], to enable systems with adaptability to unknown and hazardous environments [5], and to create wearable, assistive devices [6]. This emerging technology is completely revolutionising our notion of the word "robot". The potential impact of the field is enormous, it does not merely refer to a differentiation in the shore hardness of the materials from which such machines are built, instead it is used as an umbrella term for systems that are powerful demonstrations of embodied intelligence.

*B. Standardisation enables creativity and establishment of a marketplace*

Dealing with matters related to material choice, and manufacturing methods, is an inescapable reality associated with carrying out research work in this nascent, yet now maturing, field. Table 1 shows some of timeline between basic research and translation to commercial activity. It is clear that the field is maturing towards a need for standardisation to enable sharing of digital designs that have been built for a standardised process - this route is what underpins the "FabLab" philosophy; the establishment of standards and foundry processes is what enabled the microelectronics revolution.

Designers in microelectronics can be "creative within the process", safe in the knowledge that whatever they design will be able to be manufactured. Intellectual Property Blocks or "IP Blocks" enable teams to inherit designs from libraries and it enables a marketplace of licensing and sub-licensing.

The FabLab initiative enabled an open-source sharing of designs that can be manufactured on a set of standardised tools. Prototypes and products that are designed and built in any one FabLab can be reproduced in any other FabLab simply by sharing the digital files. Subscribers to the FabLab charter benefit from this standardised set of tools and processes.

Here, we want to release an initial piece of this prototyping/foundry process, and also to begin the groundwork for how the community can share these IP blocks - either open-source or under a commercial license, this is described further by Section V.

*C. The promise of robots that walk out of the machine.*

A substantial portion of the highest-impact work in fluidic soft robots from the past years has utilised 3D printing as the primary manufacturing method, largely due to its versatility, and its ability to produce complex geometries. There have been many powerful demonstrations of what is possible when using the most cutting-edge, precise machines [7], [16], [8], [9]. Unfortunately, it seems unlikely that technologies such as polyjet printers, or Digital Light Processing (DLP) are a solution that will enable the field to scale sustainably, this is for three main reasons: 1.) the current high cost of the printers; 2.) their extensive physical footprint, and 3.) their relatively low availability. The same factors apply to other technologies such as Stereolithography (SLA) and Selective Laser Sintering (SLS) printing, which can also come with limitations in terms of printable geometries, or which require elaborate post-processing that is often carried out in dedicated environments, due to risk of contamination of hazardous solvents or powders.

**Table 1: The path to helping soft robotics cross the translational barrier**

|  | **Stage 1:** <br> **Technology demonstration** | **Stage 2:** <br> **Early developments in desktop fabrication** | **Stage 3:** <br> **Standardised process for adoption at scale** |
|---|---|---|---|
| **Examples** | Wehner et al. (2016) [7] <br> Hubbard et al. (2021) [8] <br> Buchner et al. (2023) [9] | Zhai et al. (2023) [10] <br> Conrad et al. (2024) [11] <br> Gepner et al. (2024) [12] <br> Kendre et al. (2024) [13] <br> Zhai et al. (2025) [14] <br> Aygül et al. (2025) [15] | The **Flex Printer** platform proposed in this paper. |
| **Approximate printer costs** | In excess of $100,000 | $500 - $3,000 | <$500 |
| **Technology type** | Proprietary, cutting-edge printing methods. | Consumer-grade printers modified by individual labs, or entirely custom-built. | Open-source platform co-developed by the community. |
| **Turnaround** | Slow (parts made to order) | Rapid (in-house) | Rapid (in-house) |
| **Total print durations** | Depending on technology, post-processing requirements, and procurement times. | Hours to multiple days | Hours |
| **Level of required expertise** | Collaboration with industrial experts. | Specialised lab member know-how. | Democratised access for the entire research community. |
| **Goal** | Showcasing what is possible with the highest-precision techniques. | Validating that desktop fabrication is possible. | Making the manufacturing process a solved problem; shifting focus to delivering translational applications. |

Fused Deposition Modelling (FDM) has the upper edge in these three aspects, offering rapid in-house fabrication capabilities at an affordable price point. There have been several notable advancements in the space, and yet there are still major obstacles preventing truly scalable adoption. One of the main challenges is technical: fluidic soft robots require thin and highly elastic, yet airtight membranes. Unfortunately, the lower the shore hardness of an elastomeric material gets, the harder it is to 3D-print using FDM.

Filament extrusion systems involve pushing a thin column of thermoplastic polymer through a heated nozzle. This type of extrusion becomes highly problematic when the material is "ultra flexible"; which we define here as a term describing commercially-available thermoplastic polyurethane (TPU) filaments with a shore hardness of around 80A and below. Extruding these materials is analogous to trying to push on a piece of string: it easily buckles, and commonly results in extruder jams and inconsistent extrusion. The print defects that result from these issues are particularly problematic for fluidic systems that require near-perfect airtightness to operate reliably [12], [17]. Despite a long history of promises of the idea of FDM printing soft actuators (e.g. work by Yap et al. [18] in 2016), notable advancements in printing more advanced systems have only been more recent. Some, like Kendre et al. [17], [13] have made significant progress in printing elastic components for fluidic logic architectures, but the end systems still require additional rigid parts, and manual post-processing assembly.

Being able to produce entire monolithic soft robots rapidly and reliably with integrated fluidic control, actuation, and sensing would not only enable new types of innovations in the research field, but also unlock translational applications. The fact that these machines present no spark risk, and are unperturbed by ionising radiation or high-magnetic fields will enable innovation in high-explosion risk areas (oil and gas), nuclear decommissioning sectors and space exploration, as well as biomedical devices that can operate inside an MRI machine. Such robots could be printed and deployed directly at the point of use in these environments.

There are two published examples of fludic walking robots that have been printed monolithically [11], [14] (shown in Figure 1 b), but until now no one had been able to achieve the "holy grail": printing a robot which autonomously walks out of the machine which made it. In this paper, we demonstrate how we managed to reach this important milestone. Moreover, we introduce a framework within which others will be able to easily build upon this result, and make it much more than a one-off achievement. This robot has a bill of materials comprising of only one line - flexible TPU filament - the implications of creating machines of this type are profound as they have the following design for manufacturability characteristics: supply chain resilience, reduction in the number of failure points from sub-systems, and the potential to be decomposed and re-formed into new filament enabling a truly circular economy.

*D. The Flex Printer is an open-source, low-cost platform which aims to facilitate the next wave of scientific advancements in soft robotics*

In this paper, we introduce the Flex Printer, shown in Figure 2, an open source design that can be assembled for less than $500, and solves the key reliability challenges associated with FDM printing "ultra-flexible" TPU, enabling virtually anyone to rapidly, and reliably, produce fluidic soft machines, even without prior experience. The Flex Printer can also create entirely new types of soft systems, thanks to a series of innovations described in Section II. We wish to make this a platform that will not only provide democratised, worldwide access to the technology but also help to standardise the manufacturing processes used in the field; we believe this step is necessary to truly enable soft robotics to scale.

## II. RESULTS

Here we describe five of the key hardware modifications that make up the Flex Printer. The sixth, and perhaps the most important, innovation is discussed in Section III, where we describe how printing ultra-flexible materials in the upside-down orientation enables one to print structures that were previously unfeasible.

*A. Using wider filament diameters allows to eliminate the fundamental issues of printing ultra-flexible materials.*

One of the biggest problems associated with printing flexible materials is that the column of ultra-flexible filament easily buckles under compressive load, i.e. when pushed through the heated nozzle. Trying to extrude too fast, or with jerky movements, results in inconsistent extrusion, or even excess buildup in pressure which can cause jamming of the filament in the extrusion system, or even tangling around the extruder gear, requiring tedious disassembly to fix the fault.

We were able to overcome this issue by using 2.85 mm diameter filament diameter (wider than the commonly used 1.75 mm variant). The larger cross-sectional area of the column makes it around 7 times harder to buckle (based on Euler's buckling theory, the critical buckling load $P_{\text{cr}}$ for a circular cross-section solid column $P_{\text{cr}} \propto d^4$, so the ratio for the two diameters is $\frac{P_{\text{cr1}}}{P_{\text{cr2}}} = \left(\frac{2.85}{1.75}\right)^4 \approx 7.03$), and also means that the same extruder can push it through the nozzle with more force. This single change almost eliminated the jamming issues for us. Over 12 months of around-the-clock printing on 2 machines (using Recreus FilaFlex 63A[1], which is the lowest shore hardness commercially available filament), we've experienced only three extruder jams; in neither instance did the filament wrap itself around the extruder gears. This modification significantly increases the tolerance to suboptimal print settings and toolpaths, decreasing the required level of expertise.

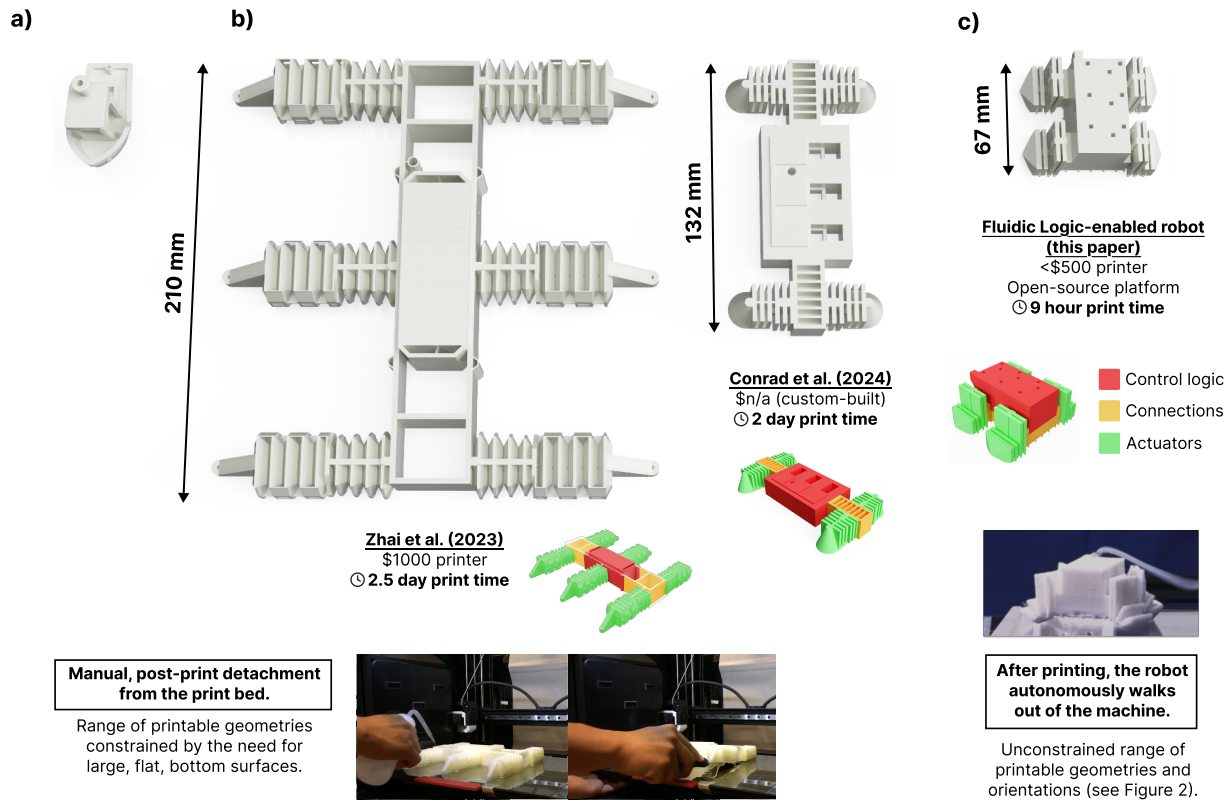

**Figure 1: We propose a new benchmark for advancements in fluidic soft systems.** a) The "3D Benchy" test print has been commonly used to assess improvements in 3D printing speed, and quality. b) A monolithically-printed, walking robot is a powerful demonstration of fluidic embodied intelligence and demonstrates the degree of integration of fluidic control systems. The key variables at play are the physical size of the robot and/or logic modules, total print duration, as well as any additional, in-built capabilities (a qualitative assessment of embodied intelligence). c) The robot we demonstrate in this paper achieves the first-time milestone of autonomously walking out of the machine that made it. Simultaneously, it exhibits a multi-fold increase in the level of system integration relative to its predecessors.

*B. Printing at very high accelerations and travel speeds minimises the need for retraction*

After finishing one segment of a toolpath and moving on to another one, FDM printers commonly retract the filament slightly to prevent "oozing" over unwanted parts of the geometry. Printing without retraction can make it easier to achieve consistent extrusion, but it distorts the dimensional accuracy of prints. More importantly, it can cause blockages of fluidic channels, compromising the print reliability of soft robotic architectures. [10] suggested modifying the geometry and slicing profiles to facilitate continuous "Eulerian" toolpaths that minimise the need for retraction. Whilst we don't think this is strictly necessary (we successfully used retraction for all of our prints in [12]), the use of retraction does require careful tuning, steepening the learning curve for new users.

The Flex Printer allows to avoid any retraction tuning (without having to compromise on geometrical features) due to printing at very high accelerations (upwards of $10000 \frac{\text{mm}^2}{s}$) and travel speeds (upwards of $500 \frac{\text{mm}}{s}$). When moving so fast between positions, the filament ooze is minimised, and retraction can be turned off completely. This capability is enabled by the CoreXY motion system, resonance compensation (Klipper [https://github.com/Klipper3d/klipper "input-shaping"]), as well as the compact (42x23x23cm), rigid frame of the Voron0 open-source design[2] on which the printer is based. The smaller 12x12x12cm build volume is an adequate compromise, given the current focus on smaller-scale fluidic systems. A small print bed also minimises the need for accurate bed leveling.

---

[1] https://recreus.com/gb/filaments/1-6-filaflex-60a.html

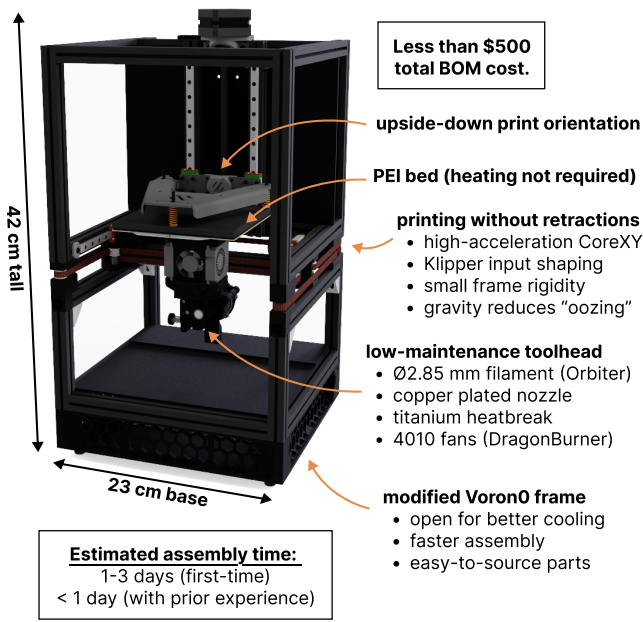

**Figure 2: The key features of the Flex Printer.** Its overarching design philosophy centres on removing unnecessary parts, and introducing modifications that enormously reduce the need for regular maintenance. The printer is optimised to be as easy as possible to build and operate, even by newcomers to the field. For an in-depth explanation of the individual hardware features, refer to Section II. In Section III, we describe the possibilities granted through printing in the upside-down orientation.

*C. Excellent adhesion to PEI eliminates the need for a heated bed*

The Flex Printer features a polyethemiride (PEI) sheet, which has excellent adhesion characteristics with TPU. The adhesion is so good, in fact, that the use of a heated bed is no longer required, even when printing upside down (the advantages of which we describe in Section III). Removing the heated bed significantly simplifies the assembly procedure, and reduces the overall bill of materials (BOM) cost.

*D. Open design for better cooling*

Better cooling not only allows to print faster and more reliably but also helps with bridging and overhangs, making complex geometries printable without supports. We use the DragonBurner extruder[3] with large 4010 fans to achieve better cooling. Our design does not use an enclosure; this open design not only improves air circulation but also simplifies the assembly process.

*E. Extruder components that minimise the need for regular maintenance*

Copper plated nozzles have good thermal transfer characteristics, and TPU does not stick to them it as easily as to typical brass nozzles. This quality minimises the likelihood of filament jams, and reduces the need for regular maintenance.

The hotend features a titanium alloy heatbreak which dissipates heat better than its stainless steel counterpart. This characteristic prevents premature heating of the filament outside of the melt zone. We use the open-source, direct-drive Orbiter F2.85 extruder[4] it is cheap and easy to source, yet its orbital gearbox generates ample amounts of force, making it easier to achieve consistent extrusion. The filament path is also reasonably constrained, minimising the likelihood of filament tangling itself into the extruder gears.

### III. Upside-down FDM printing enables to rapidly prototype previously unprintable, ultra-flexible structures.

FDM printing in the upside-down orientation has been a niche topic recently popularised by the open-source Positron project[5]. What had not been done before this paper is exploring this mode of printing in the context of ultra-flexible materials. We have discovered that printing upside-down introduces revolutionary new capabilities for the development of fluidic soft systems, which we illustrate in Figure 3.

*A. Wide, leaktight bridging surfaces are now possible.*

Printing long unsupported geometries is referred to as "bridging". When printing with regular filaments, optimal cooling and calibration usually allows the creation of high-quality bridges, i.e. without "sagging" with gravity. Ultra-flexible filaments are particularly challenging in this regard due to their elasticity; extreme sagging is the norm, and the drooping strands of filament don't fuse with each other, making it difficult to create leak-tight horizontal membranes for fluidic systems.

As discussed in Section II.B, minimising channel blockages is key to the successful printing of fluidic soft systems, so any sagging should be minimised, if not eliminated completely. Reducing this problem can be achieved with appropriate design for manufacture (i.e. minimising unsupported horizontal structures), but such an approach constrains the range of printable geometries.

---

[2]https://github.com/VoronDesign/Voron-0
[3]https://github.com/chirpy2605/voron/tree/main/V0/Dragon_Burner
[4]https://www.orbiterprojects.com/orbiter-f2-85/
[5]https://www.positron3d.com/

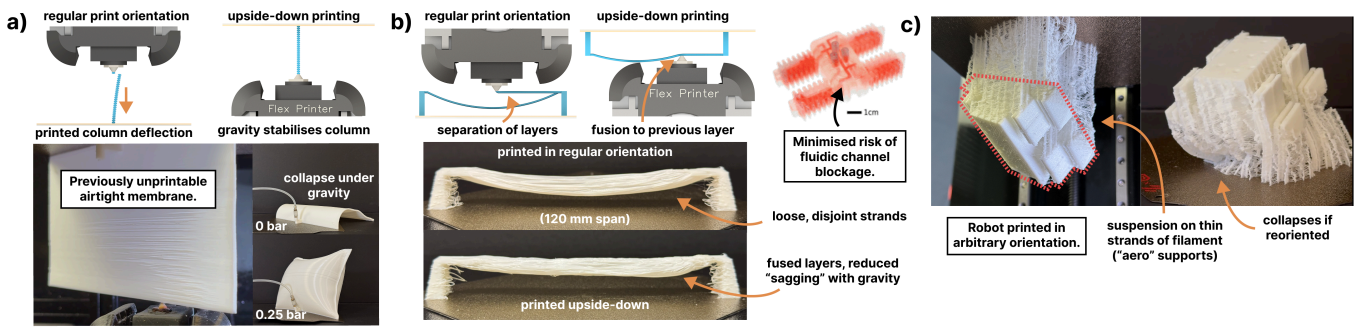

Figure 3: **The Flex Printer makes it possible to create entirely new types of soft systems.** a) Printing in the upside-down orientation allows one to print thin, airtight membranes reliably even those of arbitrary heights (the pictured membrane is 118 mm tall, filling the entire printable volume). Airtight membranes are essential for the operation of fluidic systems. b) Avoiding blockages in fluidic channels significantly improves the reliability of monolithically-printed fluidic architectures. Printing upside-down also significantly improves the airtightness of horizontal, thin membranes. c) Thin column "aero" supports enable to print of virtually any geometry in any print orientation, as if "suspended in air". Similar types of structures could also be used to integrate new metamaterial geometries into soft systems.

We discovered that when printing in the upside-down configuration, the issue is almost fully eliminated (as shown in Figure 3 b). The Flex Printer is able to bridge significantly longer gaps thanks to this modification; this ability also contributes to greater consistency in printing channels for fluidic architectures.

### B. There is a reduced need for supports

In the upside-down orientation, it is also significantly easier to print thin vertical membranes, as gravity now helps to stabilise the vertical structure (tensile load), rather than causing it to buckle (compressive load) under its own weight. This characteristic is of profound importance for printing soft systems and robots, which commonly require thin deforming membranes to operate. Figure 3 a) shows an example of a structure that would have been unprintable in the regular orientation.

### C. Thin "aero supports" enable complex print geometries which are "suspended in air".

Thanks to the tensile loading characteristic described in Section III.B it is now also possible to print new types of support structures – an array of thin vertical columns (one to three nozzle extrusion widths wide) that collectively support the weight of the printed object, but are very easy to remove after printing – something which can be problematic with conventional TPU support structures (which can easily fuse with the object). We call these structures "aero supports" to highlight their light, minimalist footprint which not only aids in their removal but also reduces the total print time and filament usage.

Another interesting characteristic of these supports is that they constrain the object almost exclusively in the axial (vertical) direction. In Section IV, we demonstrate how we utilised this characteristic to print the first robot to autonomously walk off the print bed, right after printing. This feat had not been possible before, given that large flat surfaces situated on the bed were required to make structures printable.

## IV. Demonstrating a robot walking off of the 3D printer bed

### A. Fluidic robot design, optimised for additive manufacture

Figure 4 illustrates the CAD design of the robot, along with its supporting structures. The gait of the robot utilises the design by Conrad et al. [11], featuring two "ligament" actuators to move each limb laterally, and one "foot" actuator to lift the limb off the ground.

To ensure that the robot would be capable of walking off of the bed, we needed to minimise its size and mass footprint. We were able to accomplish this by utilising a highly integrated Fluidic Logic architecture; its geometry and routing is automatically generated and optimised for the 3D printing method. At the core of the robot is a CMOS pneumatic ring oscillator which outputs a 3-phase oscillating pressure signal.

### B. Test procedure

In order to operate the robot we connected it (post-print) to a pneumatic pressure source providing positive pressure of 2.25 Bar. When making the connection we took special care not to accidentally dislodge the printed supports, as that would constitute cheating in our aim of making the robot walk off the print bed. The robot had to remain hanging off of the build plate after connecting the pressure supply.

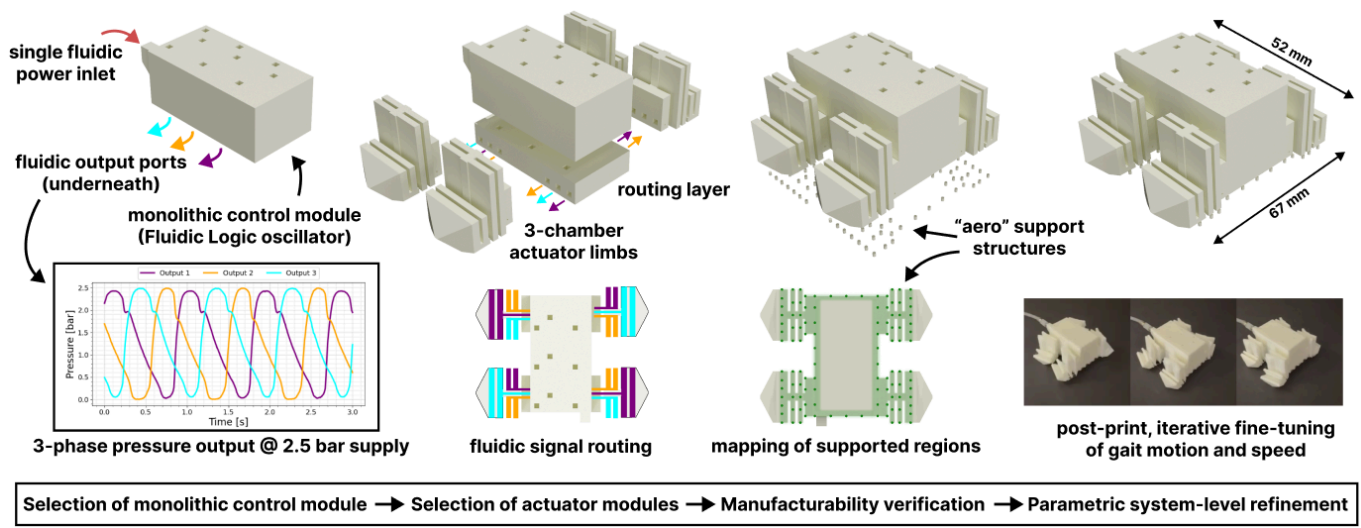

**Figure 4: An integrated systems approach to creating robots with fluidic embodied intelligence.** We build soft systems from libraries of standardised component building blocks, which come pre-optimised for manufacturability. The fluidic timing of oscillator components changes drastically after connecting any end-effectors, connectors, or sensors, so refinement needs to be carried out at the system level. Due to the rapid prototyping speed of the Flex platform, experimental characterisation is a feasible option, even in the absence of a precise analytical, or simulation model. This process is analogous to manufacturing processes utilised in the electronics industry, where, due to the complexity of the underlying physical phenomena, the refinement of models through experimental verification plays a crucial role.

The printed robot is upside down, so we designed a mechanism that automatically detached the build plate after the print was finished and reoriented the build-plate plus robot into an upright position for the walking test. Figure 5 shows the key still-frames from this entire procedure; see Supplementary Video 1 for the full recording.

## V. Fostering collaborative development of the open-source platform

### A. How to become a contributor

We have decided to distribute ownership of the platform, hosting the printer files on "neutral ground": a Github repository[6] administrated through the Soft Robotics Forum, an online platform hosted by the IEEE Technical Committee for Soft Robotics; we hope that this will encourage others to contribute to the project, as something that belongs to the whole community rather than one particular research group. We have also set up a dedicated channel on the Soft Robotics Forum server on Discord[7] this online space is aimed at facilitating collaborative research between members of the international research community. The dedicated channel will enable newcomers to seek help and ask questions when assembling the printer, and then later contribute to the project by offering their own advice, as well as ideas and suggestions for improvements. All custom modifications can also be directly submitted as pull requests to the shared repository, which also includes detailed instructions for how to do so.

### B. Directions for future development

We invite others to actively contribute to the project, and help shape it into a platform that will offer democratised access to this technology to both researchers, as well as students, around the whole world. Some of our proposed suggestions for the next steps include implementing closed-loop printing and calibration methods such as those discussed by Wu et al. [19] and Read et al. [20], multi-material printing capabilities such as these discussed by Aygül et al. [15] as well as a re-design of the extruder to further improve cooling, shorten the filament path, and automate any occasional maintenance procedures (the full list of ideas is available on the Github page, and will always remain open to new suggestions). Instead of competing to solve the same problems, we hope that the introduction of the Flex Printer platform will help make FDM printing of soft robots a solved problem, and that it will empower the whole community to focus on making the next wave of breakthrough advancements in soft robotics.

---

[6]https://github.com/The-Soft-Robotics-Forum/flex-printer
[7]Join link: http://www.softrobotics.forum

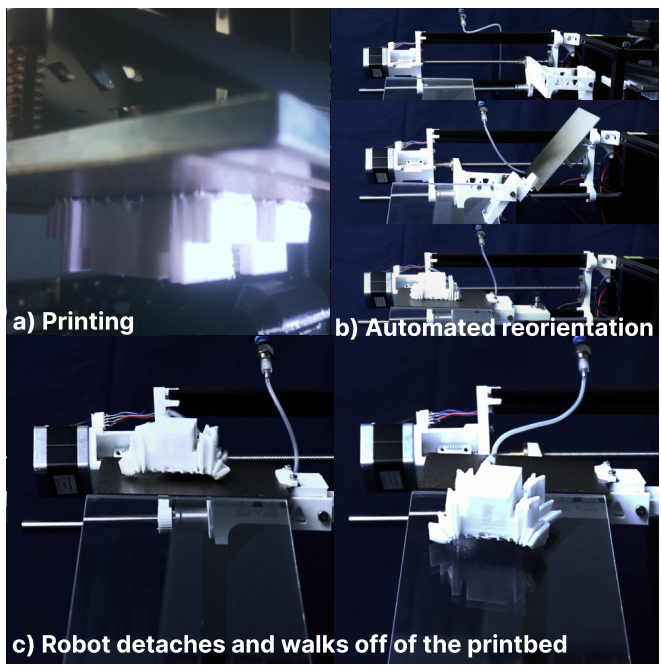

**Figure 5: We demonstrate a robot which autonomously walks off of the print bed.** a) The monolithic robot is printed in an 8-hour, 52-minute-long single run. b) After printing, the fluidic power port is carefully connected, and the bed is removed, and reoriented using an automated mechanism. c) The robot is powered by a supply pressure of 2.25 bar, and walks off the print bed. See Supplementary Video 1 for the entire procedure.

*C. The Bestiary of Fluidic Machines, an online repository of reconfigurable fluidic building blocks*

To help lay the foundation for the new economy emerging around the development of reusable fluidic system blocks (which we have discussed in Section I), we have also set up another repository for sharing fluidic system IP blocks with the rest of the community[8]. This resource makes it easy to reconfigure existing fluidic component blocks into entirely new designs, for example, the designs introduced in this paper. We encourage others to submit their design blocks to the platform and help turn it into a thriving marketplace of ideas that will rapidly accelerate the pace of development of fluidic soft systems.

## VI. Conclusions

In this paper, we introduced the Flex Printer project, featuring an open-source FDM 3D printer that aims to solve the fundamental manufacturing difficulties that are holding back progress in printable soft robotics. We proposed a set of hardware modifications that allowed us to overcome fundamental issues related to the printing speed, reliability, and ease of use. We demonstrated how the printer allows to reliably print ultra-flexible materials without requiring extensive, specialised knowledge.

We also showcased how our new proposed method of upside-down printing of ultra-flexible filaments is ideally suited to printing fluidic control architectures and soft robots, which require thin vertical membranes, and complex 3D geometries. To demonstrate the capabilities of this technology, we showcased how we used it to 3D print a soft robot that was able to autonomously walk off of the 3D printer bed right after it was printed, a first-time milestone in this field.

## VII. Data Availability

All of the files related to the Flex Printer are available on the official Github repository (see Section V). The video of the walking robot is in Supplementary Video 1, and using the link[9]. Uncut timelapse recordings of the print are available upon request. The CAD files of the Fluidic Logic oscillator are freely available on the Fluidic Machine Bestiary (see Section V.C).

## VIII. Conflict of Interest

The authors are associated with Fluidic Logic Ltd., a stealth startup registered in Scotland (SC688151).

## IX. Author Contributions

M.G: Conceptualization, methodology, and writing. J.M: Methodology and writing. A.A.S: Writing, review, and editing.

## X. Funding

This research was funded by EPSRC Centre for Doctoral Training in Robotics and Autonomous Systems (CDT-RAS), grant number EP/S023208/1.

---

[8]https://github.com/The-Soft-Robotics-Forum/fluidic-machine-bestiary

[9]https://vimeo.com/1056893158/3d4e62b8bc